# Reliability-Aware Ensemble Classification Under Class Imbalance: A Calibration Study on Liquid-Based Cervical Cytology

Nisreen Albzour, Sarah S. Lam

*School of Systems Science and Industrial Engineering, Binghamton University, Binghamton, NY 13902, USA*

* Corresponding author: nalbzour@binghamton.edu

ORCID: https://orcid.org/0009-0000-9317-340X

## Abstract

Cervical cytology classification models are typically evaluated on curated, class-balanced benchmarks, but real-world liquid-based cytology (LBC) collections are often small and class-imbalanced. This paper presents a class-imbalance-aware and calibration-aware ensemble classification study on the Mendeley LBC dataset, using its native four-class Bethesda taxonomy (NILM, LSIL, HSIL, SCC) rather than a collapsed binary formulation. Three lightweight architectures (Swin-Tiny, TinyViT-5M, DenseNet121) are trained directly on Mendeley LBC using weighted random sampling to counteract class imbalance, and compared against two soft-voting ensembles (Hybrid-2, Hybrid-3). Post-hoc temperature scaling is fit on a held-out calibration subset carved out of the training portion of each cross-validation fold, distinct from both the training data used to fit model weights and the evaluation fold used for final metrics, avoiding the optimistic calibration estimates that result when the same data is used for both purposes. Calibration substantially reduces expected calibration error, Brier score, and negative log-likelihood for every model and ensemble configuration tested, while discrimination metrics (accuracy, macro-F1, macro-AUROC) remain essentially unchanged. Ensemble size shows no consistent additional reliability benefit over the best individual model once all configurations are properly calibrated. Confusion matrices show that all classification errors, across every configuration, are confined to the boundary between high-grade lesions (HSIL) and carcinoma (SCC); no errors involve the negative (NILM) or low-grade (LSIL) categories. These results suggest that, for this dataset, calibration is the dominant lever for reliability, not ensemble size, though this conclusion should be read in light of the dataset's modest size.

## 1. Introduction

Automated classification of cervical cytology images has been studied extensively on curated benchmarks such as SIPaKMeD, where class representation is close to balanced by design. Real-world liquid-based cytology (LBC) collections rarely share this property: class frequencies reflect actual screening populations, in which high-grade and malignant findings are, by definition, rarer than negative findings. In one reported LBC screening program, for example, the negative (NILM) category accounted for roughly 89% of 1,293 samples while high-grade lesions accounted for well under one percent [1]. Models evaluated only on balanced benchmarks provide limited evidence about how they will behave when trained and deployed on this kind of skewed, real-world data.

This paper studies the Mendeley LBC dataset directly, rather than as an external validation target for models trained elsewhere. Two aspects of this dataset motivate the specific focus of this study. First, its class distribution is naturally imbalanced, dominated by the negative (NILM) category, which requires explicit imbalance handling during training rather than accuracy-only evaluation. Second, prior work evaluating models on Mendeley LBC as an external-validation set has typically collapsed its native four-class Bethesda taxonomy into a binary Normal-versus-Abnormal task to reconcile it with a differently-labeled training source. Binary formulations are common in LBC deep learning more generally; whole-slide LBC screening models, for instance, are frequently trained to separate neoplastic from non-neoplastic specimens rather than to grade lesion severity [2]. That collapse discards clinically relevant distinctions between LSIL, HSIL, and SCC. Here, the native four-class taxonomy is retained throughout, and all models are trained directly on Mendeley LBC rather than transferred from a different dataset.

The contribution of this paper is a focused empirical study of two reliability levers, class-imbalance handling via weighted sampling and post-hoc calibration via temperature scaling, on this dataset, together with a comparison of individual models against small soft-voting ensembles (Hybrid-2, Hybrid-3). Consistent with prior work showing that neural networks are often poorly calibrated even when accurate, calibration is evaluated separately from discrimination throughout, using a calibration subset that is held out from both model training and final evaluation to avoid overstating calibration quality.

## 2. Related Work

This section reviews the four bodies of work that bear on the study: the Mendeley LBC dataset and deep learning for cervical cytology classification (Section 2.1), ensemble methods applied to this domain (Section 2.2), calibration and uncertainty estimation in medical imaging (Section 2.3), and class-imbalance handling (Section 2.4). Each has been studied largely in isolation. Cytology classification and ensemble work report discrimination metrics and rarely examine whether predicted probabilities are trustworthy; the calibration literature is developed mostly on other imaging modalities; and imbalance handling is treated as a training-time concern separate from the reliability of the resulting probabilities. The gap this paper addresses lies at their intersection.

### 2.1 Dataset and Cervical Cytology Classification

The Mendeley LBC dataset was introduced as a liquid-based cytology resource annotated under the four-class Bethesda taxonomy (NILM, LSIL, HSIL, SCC) that this paper retains [3]. The present study uses the 962 images obtained after preprocessing and loading, as reported in Section 3.1. It sits within a broader set of cervical cytology benchmarks: DCCL provides a substantially larger Bethesda-labeled resource, comprising 14,432 patches drawn from 1,167 whole-slide images [4], which illustrates the difference in scale between whole-slide collections and the smaller image-level datasets on which most published classification results are reported. Deep transfer learning is the standard approach to classification on such data. Wong et al. report Bethesda-class grading of liquid-based Pap smear images using ResNet50V2 and ResNet101V2 backbones, with high reported precision on the HSIL and SCC categories [5]. More recently, Coskun et al. combine three public repositories with a proprietary multi-center whole-slide cohort to study scanner-induced domain shift in cell-level LBC classification, harmonizing labels into four Bethesda categories (NILM, ASC-US, LSIL, HSIL) [6]; their taxonomy differs from the one used here in that it includes ASC-US in place of SCC, so their results are not directly comparable to ours, but their finding that single-source training limits robustness is relevant context for a study, such as this one, trained on one public source. None of these studies report calibration metrics. Our own earlier segmentation-plus-classification pipeline for Pap smear images [7] shares this limitation; it was evaluated on a different public benchmark and reported discrimination only, and is cited here as a related methodological example rather than as evidence about Mendeley LBC.

### 2.2 Ensemble Learning in Cervical Cytology

Ensemble methods, most commonly implemented via soft voting over independently trained models, have been applied to cervical cytology classification primarily as an accuracy-boosting strategy. Deep-learning ensembles have been proposed to assist cytopathologists in Pap test image classification [8], and fuzzy-rank and uncertainty-aware ensembling schemes have both been developed specifically for Pap smear classification [9], [10]. All three are evaluated on accuracy and F1 rather than on calibration quality. Comparatively little attention has been paid to how ensemble size interacts with calibration specifically, which is the focus of this paper.

### 2.3 Calibration and Uncertainty in Medical Imaging

Confidence calibration is widely studied as an axis of model quality distinct from discrimination accuracy, with temperature scaling established as a simple and effective post-hoc correction that rescales predicted confidence without altering predicted class rankings [11]. Within medical imaging specifically, Carse et al. compare temperature scaling against alternative calibration objectives on dermatology and histopathology datasets, providing direct empirical support for post-hoc temperature scaling on medical images [12]. Penso et al. further show that calibration remains workable under label noise by incorporating estimated label-noise structure into the calibration procedure [13], which is relevant given that cytology annotations are subject to inter-observer variability. Neither study addresses cervical cytology, and neither examines the interaction between calibration and ensemble size. Related work has since examined whether calibration holds under cross-cohort distribution shift: post-hoc calibration methods have been compared under external validation across the SEER and METABRIC breast cancer cohorts [14], and the transferability of calibration has been found to decouple from that of feature-level explanations under such shift [15]. Efforts to automate medical-image classification pipelines end-to-end, as in leukemia detection, similarly motivate reliability-aware evaluation rather than accuracy-only reporting [16].

### 2.4 Class Imbalance Handling

Class imbalance is a long-studied and general challenge in machine learning, with synthetic oversampling [17] and weighted resampling [18] among the most widely used mitigations. This paper applies weighted random sampling specifically because it requires no synthetic image generation, a relevant practical consideration for a dataset of this size. Notably, the cervical cytology studies surveyed above are evaluated on curated collections in which class representation is close to balanced by design, or on collapsed binary tasks; the combination of native four-class imbalance, explicit imbalance handling, and calibration evaluation is not, to our knowledge, treated together in prior work on this dataset. This paper combines these three considerations, class imbalance handling, calibration, and ensemble size, within a single controlled comparison on one real-world LBC dataset trained directly, rather than studying any one of them in isolation or evaluating it only as an external-validation afterthought.

## 3. Methodology

All experiments were conducted directly on the Mendeley LBC dataset. No models or checkpoints were transferred from other cervical cytology datasets; every model reported here was trained from ImageNet-pretrained weights directly on Mendeley LBC. Figure 1 summarizes the full experimental pipeline; the remainder of this section describes each stage in turn.

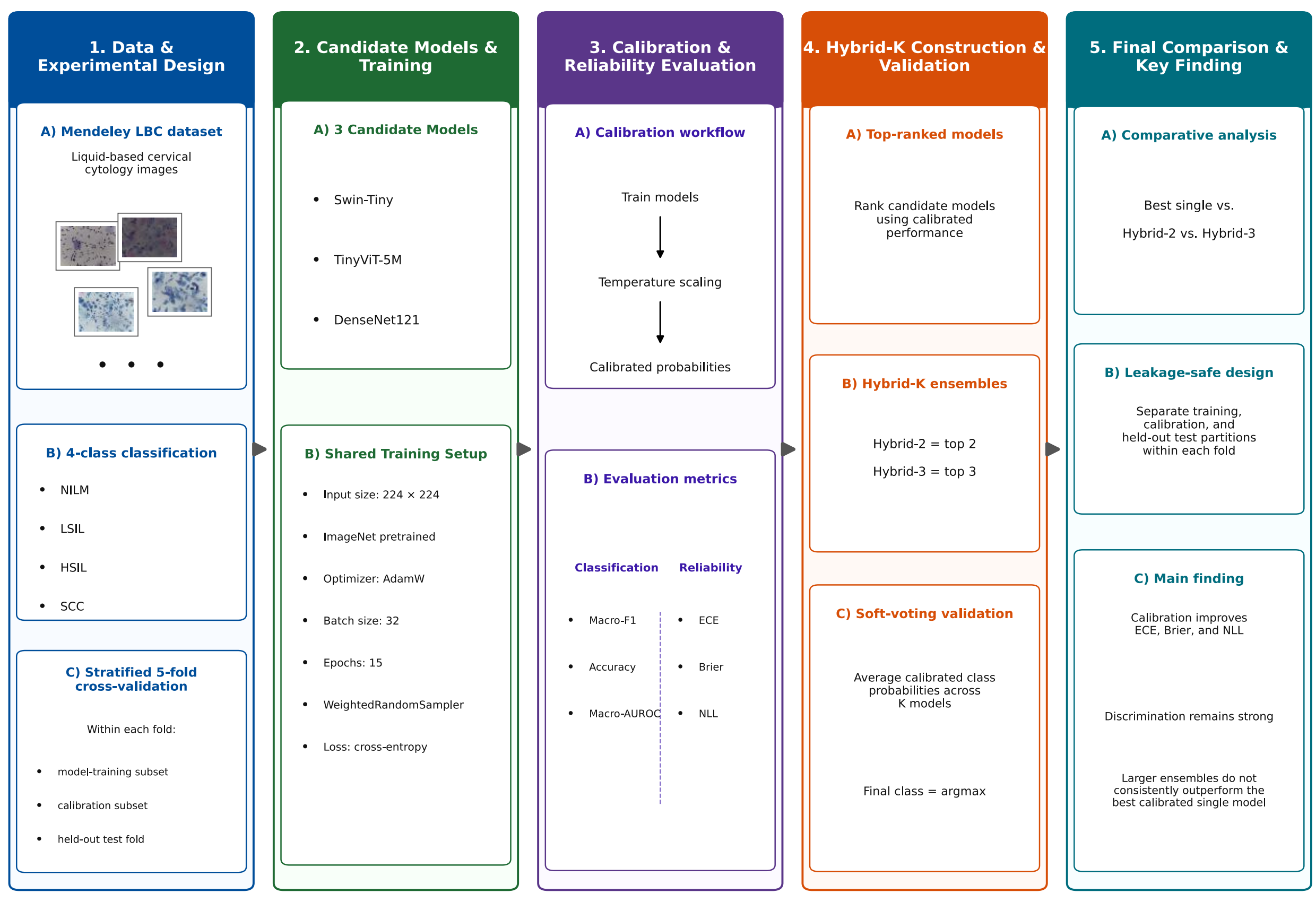

***Figure 1. Overview of the experimental pipeline: data and cross-validation design, the three candidate models and shared training setup, the calibration and reliability evaluation stage, Hybrid-K ensemble construction by soft voting, and the final comparison of the best calibrated single model against Hybrid-2 and Hybrid-3.***

## 3.1 Dataset

Mendeley LBC comprises 962 liquid-based cytology images labeled under the four-class Bethesda System: Negative for Intraepithelial Malignancy (NILM), Low-grade Squamous Intraepithelial Lesion (LSIL), High-grade Squamous Intraepithelial Lesion (HSIL), and Squamous Cell Carcinoma (SCC). Table 1 reports the class distribution and the corresponding inverse-frequency sampling weights used to counteract this imbalance during training (Section 3.4), which is naturally skewed toward the negative category. Figure 2 shows one representative image from each of the four classes.

***Table 1. Class distribution and sampling weights.***

| Class | Count | Sampling Weight |
|---|---|---|
| NILM | 612 | 0.00163 |
| LSIL | 113 | 0.00885 |
| HSIL | 163 | 0.00613 |
| SCC | 74 | 0.01351 |

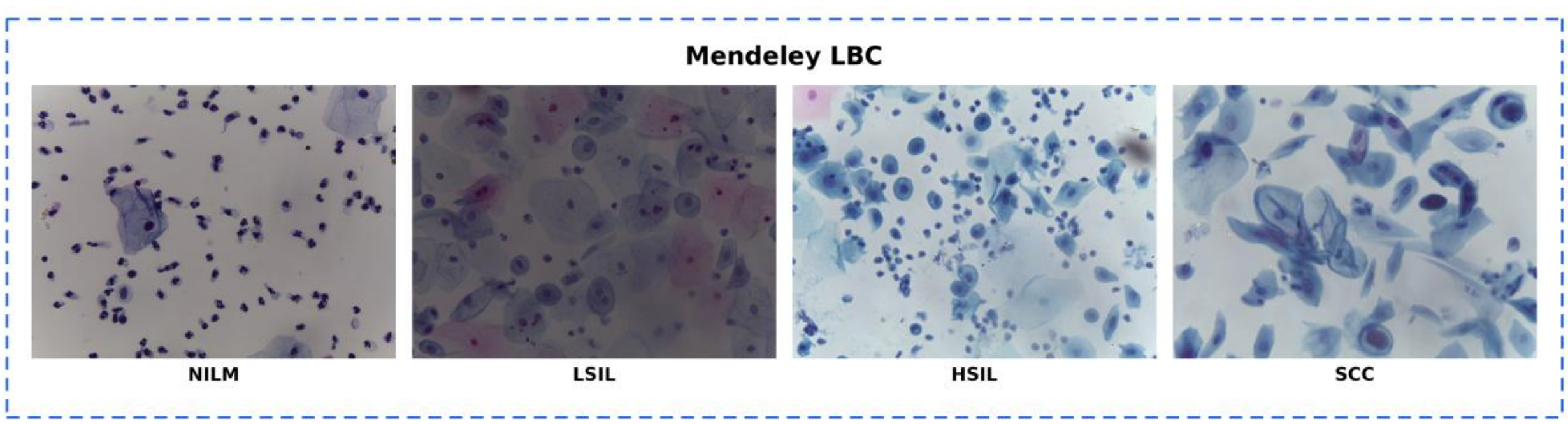

***Figure 2. Representative Mendeley LBC images for the four Bethesda classes, shown left to right in order of increasing severity: NILM, LSIL, HSIL, and SCC. One image is shown per class; each panel is a full field of view as provided in the dataset, downsampled for display.***

### 3.2 Class Formulation

The native four-class Bethesda taxonomy is used throughout; no classes are merged or collapsed to a binary task. This preserves the clinical distinctions between low-grade, high-grade, and malignant findings that a binary Normal/Abnormal formulation would discard.

### 3.3 Candidate Models and Ensembles

Three lightweight architectures were selected as a small, focused candidate pool: Swin-Tiny, TinyViT-5M, and DenseNet121. Transfer learning from ImageNet-pretrained backbones is the established approach for cervical cytology classification at this data scale [5], and lightweight transformer variants have been reported to remain competitive with heavier convolutional baselines while reducing deployment cost [19]. A reproducible benchmark of ViT-Tiny against CNN baselines on cervical cell images supports this choice of candidate pool [20]; both of those studies addressed discrimination and efficiency only, were evaluated on benchmarks other than Mendeley LBC, and did not report calibration, so they motivate the architecture selection rather than any reliability claim made here. Each model was fine-tuned directly on Mendeley LBC from ImageNet-pretrained weights. Two soft-voting ensembles were then constructed from the ranked candidates: Hybrid-2 (the two highest-ranked models) and Hybrid-3 (all three models).

### 3.4 Class-Imbalance Handling

Class imbalance was addressed during training using a WeightedRandomSampler with inverse class-frequency sampling weights, so that minority classes (particularly SCC) are sampled more frequently per epoch than their raw frequency in the dataset. This changes each sample's sampling probability during training; it does not alter the underlying dataset or its true class distribution, which is reported separately in Table 1 alongside the sampling weights it implies.

### 3.5 Calibration Protocol

Post-hoc temperature scaling was applied to calibrate each model's predicted probabilities. To avoid the optimistic calibration estimates that result when the same data is used both to fit a temperature and to evaluate it, each cross-validation fold's training portion was further split into a model-training subset (85%) and a calibration subset (15%, stratified by class), with the calibration subset held out from model training entirely. The temperature was fit by minimizing negative log-likelihood on the calibration subset only. Final metrics, both before and after calibration, are computed exclusively on the fold's held-out test partition, which is never used for either training or calibration.

As a specific leakage check: within every cross-validation fold, the temperature-scaling parameter is fit exclusively on the calibration subset described above, and every accuracy, F1, AUROC, ECE, Brier, and NLL value reported in this paper is computed exclusively on that fold's held-out test partition. No sample used to fit a temperature is ever also used to evaluate it, and no sample used for either training or calibration is ever included in a reported test metric. This partitioning discipline is consistent with leakage-aware benchmarking protocols used in related medical-imaging classification work [21]; the underlying concern, that overlapping partitions inflate reported reliability, is domain-independent, though the classification task there is unrelated to cervical cytology.

### 3.6 Cross-Validation and Evaluation Metrics

Stratified five-fold cross-validation was used, with training, calibration, and test partitions constructed as described above within each fold. Six metrics were computed on the held-out test partition of every fold: accuracy, macro-averaged F1, macro-averaged AUROC (one-vs-rest), expected calibration error (ECE), Brier score, and negative log-likelihood (NLL). Each metric is reported both before calibration (raw softmax outputs) and after calibration (temperature-scaled outputs), to isolate the effect of calibration from the effect of model or ensemble choice.

## 4. Results

Table 2 reports before- and after-calibration metrics for the best individual model (Swin-Tiny) and the two ensembles (Hybrid-2, Hybrid-3), averaged across the five cross-validation folds.

***Table 2. Best single model (Swin-Tiny) and Hybrid-K ensembles, before and after temperature-scaling calibration, averaged across 5 cross-validation folds. All metrics computed on held-out test partitions only.***

| Config | Acc. (before) | Acc. (after) | Macro-F1 (before) | Macro-F1 (after) | AUROC (before) | AUROC (after) |
|---|---|---|---|---|---|---|
| Swin-Tiny | 0.9969 | 0.9969 | 0.9928 | 0.9928 | 0.9969 | 0.9968 |
| Hybrid-2 | 0.9948 | 0.9948 | 0.9883 | 0.9883 | 0.9997 | 0.9998 |
| Hybrid-3 | 0.9948 | 0.9969 | 0.9879 | 0.9928 | 0.9993 | 0.9994 |

***Table 3. Reliability metrics, before and after calibration (same configurations as Table 2).***

| Config | ECE (before) | ECE (after) | Brier (before) | Brier (after) | NLL (before) | NLL (after) |
|---|---|---|---|---|---|---|
| Swin-Tiny | 0.0726 | 0.0037 | 0.0131 | 0.0062 | 0.0848 | 0.0306 |
| Hybrid-2 | 0.0804 | 0.0041 | 0.0167 | 0.0086 | 0.0950 | 0.0294 |
| Hybrid-3 | 0.0840 | 0.0082 | 0.0186 | 0.0083 | 0.0989 | 0.0278 |

Discrimination metrics were already high before calibration and changed little afterward, as expected since temperature scaling rescales confidence without altering the ranking of predicted classes. Accuracy ranged from 0.9948 to 0.9969 and macro-F1 from 0.9879 to 0.9928 across all three configurations, both before and after calibration. Macro-AUROC was uniformly above 0.996 in all cases, with Hybrid-2 achieving the highest value (0.9998 after calibration).

Calibration had a substantial and consistent effect on all three reliability metrics. Expected calibration error fell from a pre-calibration range of 0.0726–0.0840 to a post-calibration range of 0.0037–0.0082, a reduction of roughly 88–95% depending on configuration. Brier score fell from 0.0131–0.0186 before calibration to 0.0062–0.0086 after, and negative log-likelihood fell from 0.0848–0.0989 before calibration to 0.0278–0.0306 after. These reductions were observed for every configuration tested, including the best individual model, indicating that the calibration benefit is not specific to ensembling. Figure 3 visualizes these reductions for all three configurations.

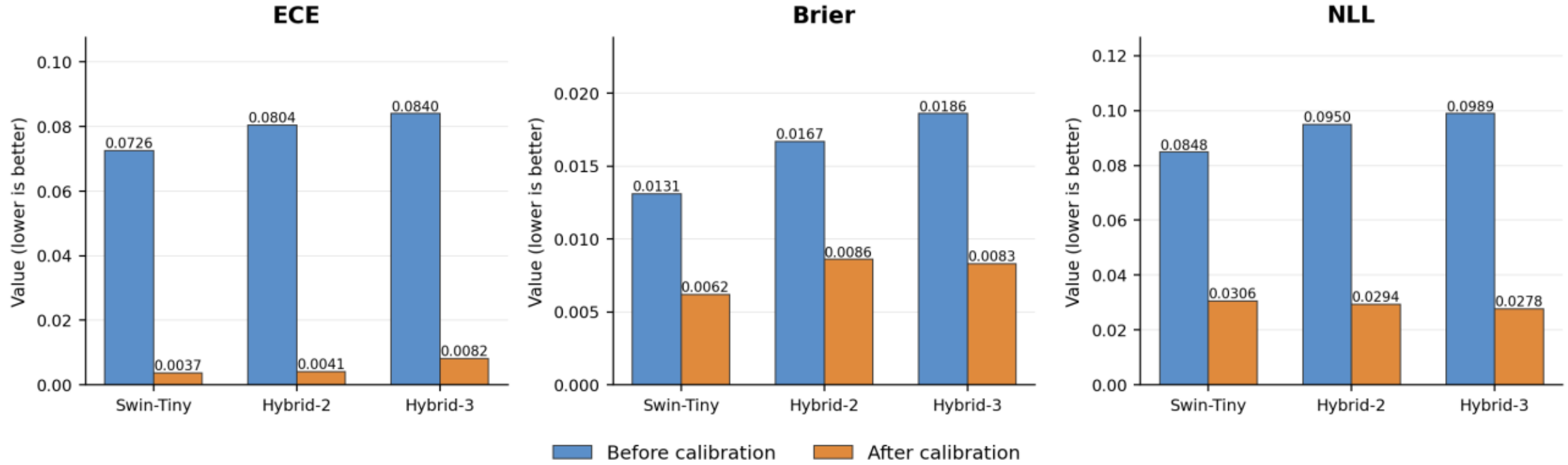

***Figure 3. Effect of post-hoc temperature scaling on reliability metrics for the best single model and Hybrid-K ensembles. Calibration substantially reduces ECE, Brier score, and NLL across all configurations, while the relative benefit of increasing ensemble size is less consistent. Values are those reported in Table 3; note that each panel uses an independent vertical scale.***

Comparing configurations after calibration, no single configuration dominated across all three reliability metrics. Swin-Tiny alone achieved the lowest ECE (0.0037) and lowest Brier score (0.0062) of the three configurations; Hybrid-2 achieved the highest macro-AUROC (0.9998); and Hybrid-3 achieved the lowest NLL (0.0278) and tied Swin-Tiny for the highest accuracy and macro-F1. No configuration was uniformly worse than the others once calibrated. This indicates that, for this dataset and this candidate pool, ensembling did not provide a consistent additional reliability benefit beyond what calibration alone already provided to the best individual model.

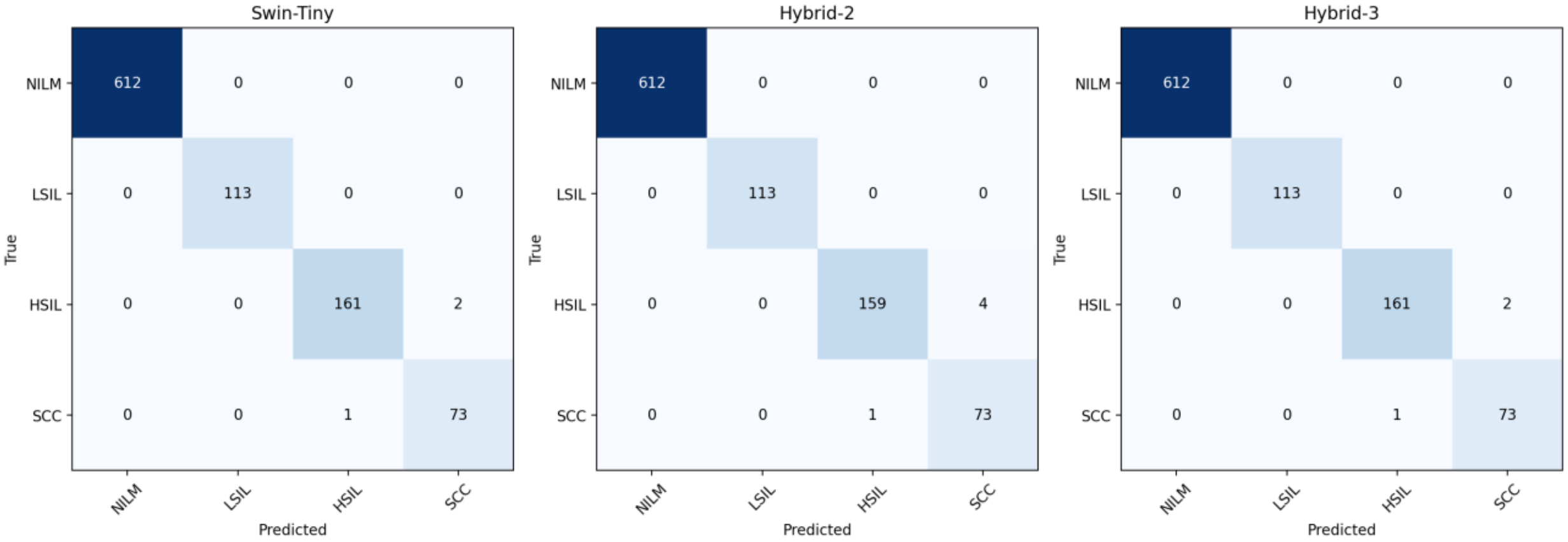

***Figure 4. Pooled confusion matrices across all 5 test folds (962 predictions) for Swin-Tiny, Hybrid-2, and Hybrid-3.***

Figure 4 shows pooled confusion matrices across all five test folds (962 predictions total) for Swin-Tiny, Hybrid-2, and Hybrid-3. All three configurations classify every NILM and LSIL sample correctly; the only errors observed, for any configuration, occur at the boundary between HSIL and SCC. Swin-Tiny and Hybrid-3 each misclassify 2 HSIL samples as SCC and 1 SCC sample as HSIL (3 errors total, 959/962 correct); Hybrid-2 misclassifies 4 HSIL samples as SCC and 1 SCC sample as HSIL (5 errors total, 957/962 correct). No configuration confuses a negative or low-grade finding with a high-grade or malignant one, or vice versa; every observed error is a confusion between the two most severe categories in the taxonomy.

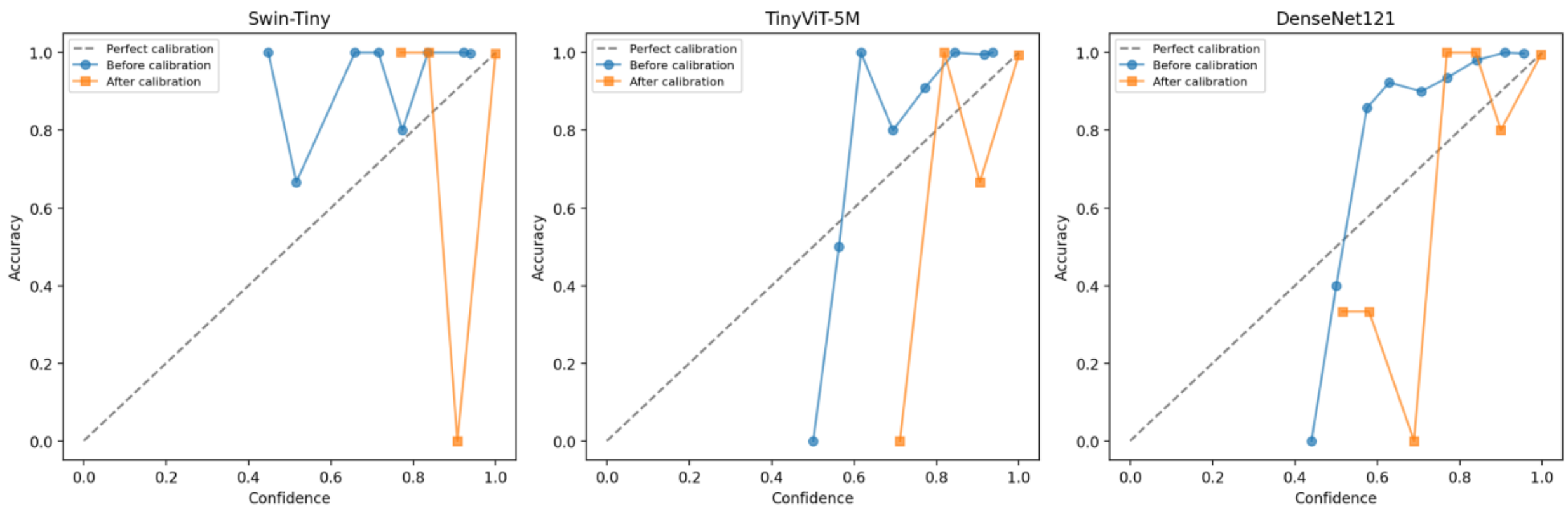

***Figure 5. Reliability diagrams (15 bins), before and after calibration, for Swin-Tiny, TinyViT-5M, and DenseNet121. The dashed line denotes perfect calibration.***

Figure 5 shows reliability diagrams for all three individual models, before and after calibration, using 15 confidence bins on the pooled held-out test predictions. The bin-level curves are visibly noisy for all three models, including several bins that swing to 0% or 100% accuracy; this is expected given that HSIL and SCC together comprise under a quarter of the dataset, so individual confidence bins can contain very few test samples and are correspondingly volatile. This bin-level noise should not be read as contradicting the strong aggregate ECE improvement reported in Table 3: ECE is a sample-weighted average across bins, so the large, well-populated bins (dominated by NILM and LSIL predictions, which are classified with near-certain confidence and are essentially always correct) drive most of the aggregate metric, while the volatile small-sample bins visible in Figure 5 contribute comparatively little to the overall ECE value despite their visual prominence in the diagram.

The accuracy and macro-F1 values reported here are very high, exceeding 0.99 in most configurations. This should be interpreted in the context of the dataset's size (962 images) and the stratified five-fold cross-validation protocol used: with four classes of substantially different size, a small number of held-out samples in the minority classes (HSIL, SCC) can produce large swings in per-fold metrics, and the near-ceiling performance observed here may not generalize to a larger or differently-composed LBC cohort without further validation. The consistency of the calibration effect across all three individually-varying models and both ensemble sizes is, for the same reason, a more robust finding than the absolute discrimination values themselves.

Table 4 disaggregates the calibration effect across the three individual architectures, reporting the fold-averaged fitted temperature alongside the three probabilistic-quality metrics before and after temperature scaling. The pattern reported for the best individual model in Table 2 holds for every architecture independently. Expected calibration error falls from 0.0726 to 0.0037 for Swin-Tiny (a 94.9% reduction), from 0.0852 to 0.0084 for TinyViT-5M (90.2%), and from 0.0830 to 0.0127 for DenseNet121 (84.7%). Brier score improves by 34.7% to 53.1% and negative log-likelihood by 38.6% to 63.9% across the same three models. Accuracy, macro-F1, and macro-AUROC are unchanged to four decimal places in all three cases, confirming that temperature scaling rescales confidence without reordering predicted classes.

*Table 4. Per-model calibration results for each individual architecture, averaged across the five cross-validation folds. T is the fitted temperature. All metrics computed on held-out test partitions only.*

| Model | T | ECE (before) | ECE (after) | Brier (before) | Brier (after) | NLL (before) | NLL (after) |
|---|---|---|---|---|---|---|---|
| Swin-Tiny | 0.250 | 0.0726 | 0.0037 | 0.0131 | 0.0062 | 0.0848 | 0.0306 |
| TinyViT-5M | 0.250 | 0.0852 | 0.0084 | 0.0234 | 0.0152 | 0.1103 | 0.0677 |
| DenseNet121 | 0.401 | 0.0830 | 0.0127 | 0.0316 | 0.0198 | 0.1143 | 0.0508 |

The fitted temperatures themselves are informative. Both transformer backbones converge to a temperature of approximately 0.25 in every fold, the lower bound of the search range, whereas DenseNet121 fits a higher and more variable temperature (fold-averaged 0.401, ranging from 0.250 to 0.533). A temperature below one sharpens rather than softens the predicted distribution, indicating that these models are under-confident rather than over-confident on this dataset, a direction opposite to the over-confidence typically reported for large networks on natural-image benchmarks. This is consistent with training under weighted random sampling, which flattens the effective class prior seen during optimization. The boundary value suggests that future work could examine a wider temperature-search range, although the present analysis keeps the pre-specified range fixed. The greater fold-to-fold variance in the DenseNet121 temperature is also consistent with the small size of the per-fold calibration subset noted in Section 3.5.

## 5. Discussion

The central finding of this study is that calibration, not ensemble size, was the dominant lever for reliability on Mendeley LBC. All three configurations were already highly discriminative before calibration; the substantial gains observed were entirely in the probabilistic-quality metrics (ECE, Brier, NLL), and these gains were realized whether or not ensembling was used. This is a materially different pattern from what might be assumed if ensembling and calibration were conflated: an ensemble is not a substitute for calibration, and calibrating a single well-chosen model can match or exceed an ensemble's reliability on a dataset of this size and composition. Table 4 strengthens this claim by showing that the calibration benefit is not an artifact of one architecture: all three backbones, spanning two transformer designs and one convolutional design, exhibit the same large ECE reduction with no discrimination cost. The fitted temperatures below one further indicate that the miscalibration on this dataset takes the form of under-confidence, so the practical benefit of temperature scaling here is sharpening well-founded predictions rather than tempering overreach. More broadly, reliability-aware evaluation is relevant across diverse engineering and pattern-recognition domains, including medical-imaging reliability analysis, speech-based identification, and reliability-oriented data-collection systems, where careful validation design, trade-off analysis, and reporting beyond aggregate performance are important [22], [23], [24].

This finding should be interpreted specifically in the context of this dataset, model pool, and calibration protocol, and not generalized beyond it without further evidence. Mendeley LBC is a comparatively small dataset (962 images across four classes), and the calibration subset carved out of each fold's training portion is correspondingly small; the stability of the fitted temperature under this data-constrained setting is itself a relevant question for future work with larger LBC collections. The choice of weighted sampling over synthetic oversampling is consistent with our earlier experience on imbalanced clinical prediction tasks [25], although that work concerned tabular clinical features rather than images, so the comparison is suggestive rather than evidential. In addition, only three candidate architectures were evaluated, and only two ensemble sizes; a broader search over ensemble composition might reveal configurations that do consistently outperform the best individual model after calibration, even if none of the three tested here did so uniformly. The confusion-matrix analysis in Figure 4 further indicates that, to the extent errors remain, future work on this dataset should focus specifically on the HSIL/SCC boundary, since no other class pair was confused under any configuration tested.

## 6. Conclusion

This paper presented a class-imbalance-aware and calibration-aware classification study on the Mendeley LBC dataset, using its native four-class Bethesda taxonomy and training all models directly on this dataset rather than transferring them from elsewhere. Temperature scaling, fit on a held-out calibration subset distinct from both training and test data, reduced expected calibration error by 85–95%, Brier score by roughly half, and negative log-likelihood by roughly two-thirds, across every model and ensemble configuration tested, while discrimination metrics remained essentially unchanged. Ensemble size did not provide a consistent additional reliability benefit once individual models were properly calibrated, and all observed classification errors were confined to the clinically difficult HSIL/SCC boundary. These results suggest that, at least for real-world LBC collections of comparable size and composition, calibration deserves at least as much attention as ensemble design when reliability, not just accuracy, is the deployment goal.